\documentclass[10pt,twocolumn,letterpaper]{article}

\usepackage{cvpr}              

\usepackage{graphicx}
\usepackage{amsmath}
\usepackage{amssymb}
\usepackage{booktabs}
\usepackage{bbding}
\usepackage{amssymb}
\usepackage{multirow}
\usepackage[accsupp]{axessibility}
\usepackage[pagebackref,breaklinks,colorlinks]{hyperref}

\usepackage[capitalize]{cleveref}
\crefname{section}{Sec.}{Secs.}
\Crefname{section}{Section}{Sections}
\Crefname{table}{Table}{Tables}
\crefname{table}{Tab.}{Tabs.}

\def\cvprPaperID{4096} 
\def\confName{CVPR}
\def\confYear{2023}

\begin{document}

\title{Delving into Shape-aware Zero-shot Semantic Segmentation}

\author{Xinyu Liu$^{1,2}$, Beiwen Tian$^{2,4}$, Zhen Wang$^3$, Rui Wang$^3$, Kehua Sheng$^3$, Bo Zhang$^3$, \\
Hao Zhao$^2$, Guyue Zhou$^2$ \\
$^1$Xidian University $^3$Didi Chuxing \\
$^2$Institute for AI Industry Research (AIR), Tsinghua University \\
$^4$Department of Computer Science and Technology, Tsinghua University \\
{\tt\small liuxinyu@stu.xidian.edu.cn}, 
{\tt\small zhaohao@air.tsinghua.edu.cn}}

\maketitle

\begin{abstract}
   Thanks to the impressive progress of large-scale vision-language pretraining, recent recognition models can classify arbitrary objects in a zero-shot and open-set manner, with a surprisingly high accuracy. However, translating this success to semantic segmentation is not trivial, because this dense prediction task requires not only accurate semantic understanding but also fine shape delineation and existing vision-language models are trained with image-level language descriptions. To bridge this gap, we pursue \textbf{shape-aware} zero-shot semantic segmentation in this study. Inspired by classical spectral methods in the image segmentation literature, we propose to leverage the eigen vectors of Laplacian matrices constructed with self-supervised pixel-wise features to promote shape-awareness. Despite that this simple and effective technique does not make use of the masks of seen classes at all, we demonstrate that it out-performs a state-of-the-art shape-aware formulation that aligns ground truth and predicted edges during training. We also delve into the performance gains achieved on different datasets using different backbones and draw several interesting and conclusive observations: the benefits of promoting shape-awareness highly relates to mask compactness and language embedding locality. Finally, our method sets new state-of-the-art performance for zero-shot semantic segmentation on both Pascal and COCO, with significant margins. Code and models will be accessed at 
   \href{https://github.com/Liuxinyv/SAZS}{SAZS}.
\end{abstract}

\begin{figure}[h]
    \centerline{
    \includegraphics[width=1\linewidth]
    {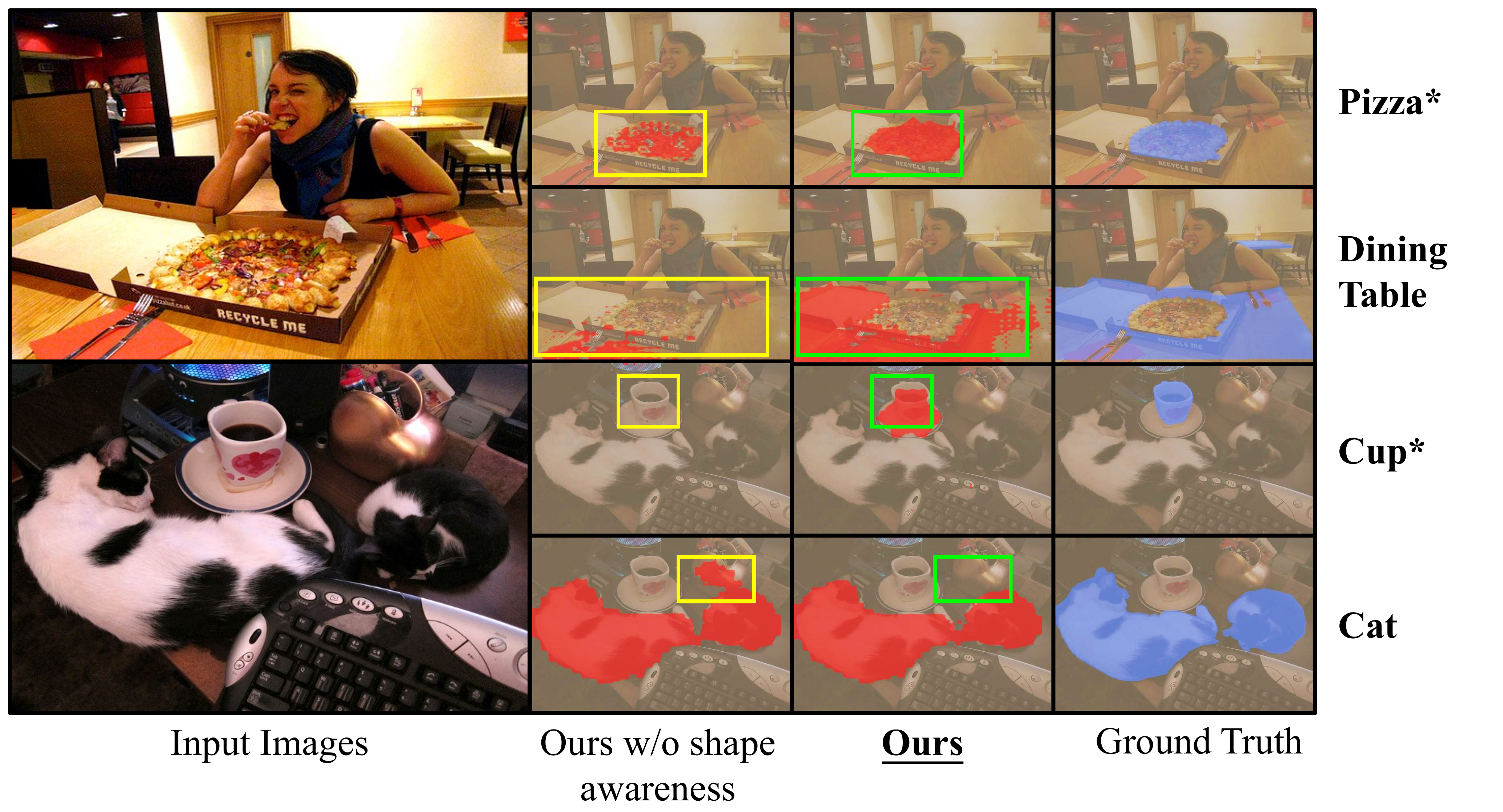}}
    \caption{
    Without retraining, SAZS is able to precisely segments both seen and unseen objects in the zero-shot setting, largely outperforming a strong baseline.
    \textbf{*} denotes unseen categories during training). }
    \label{fig:teaser}
\end{figure}

\section{Introduction}
\label{sec:intro}

Semantic segmentation has been an established research area for some time now, which aims to predict the categories of an input image in a pixel-wise manner.
In real-world applications including autonomous driving\cite{janai2020computer}, medical diagnosis\cite{liu2021review,SHEN2023106675} and robot vision and navigation\cite{9121682,chen2022cerberus}, an accurate semantic segmentation module provides a pixel-wise understanding of the input image and is crucial for subsequent tasks (like decision making or treatment selection).

Despite that significant progress has been made in the field of semantic segmentation \cite{long2015fully,chen2017deeplab,zhao2017pyramid,chen2018encoder,wang2020deep,yuan2020object,xie2021segformer,tian2022vibus,zhao2020pointly}, most existing methods focus on the closed-set setting in which dense prediction is performed on the same set of categories in training and testing time.
Thus, methods that are trained and perform well in the closed-set setting may fail when applied to the open world, as pixels of unseen objects in the open world are likely to be assigned categories that are seen during training, causing catastrophic consequences in safety-critical applications such as autonomous driving \cite{zheng2023steps}.
Straightforward solutions include fine-tuning or retraining the existing neural networks, but it is impractical to enumerate unlimited unseen categories during retraining, let along large quantities of time and efforts needed.

More recent works \cite{pham2018bayesian,bucher2019zero,gu2020context,li2020consistent,li2022language} address this issue by shifting to the zero-shot setting, in which the methods are evaluated with semantic categories that are unseen during training.
While large-scale pre-trained visual-language models such as CLIP\cite{radford2021learning} or ALIGN\cite{jia2021scaling} shed light on the potential of solving zero-shot tasks with priors contained in large-scale pre-trained model, how to perform dense prediction task in this setting is still under-explored.
One recent approach by Li et.al.\cite{li2022language} closes the gap by leveraging the shared embedding space for languages and images, but fails to effectively segment regions with fine shape delineation.
If the segmented shape of the target object is not accurate, it will be a big safety hazard in practical applications, such as in autonomous driving.

Inspired by the classical spectral methods and their intrinsic capability of enhancing shapeawareness, we propose a novel \textbf{S}hape-\textbf{A}ware \textbf{Z}ero-\textbf{S}hot semantic segmentation framework (\textbf{SAZS}) to address the task of zero-shot semantic segmentation.
Firstly, the framework enforces vision-language alignment on the training set using known categories, which exploits rich language priors in the large-scale pre-trained vision-language  model CLIP \cite{radford2021learning}.
Meanwhile, the framework also jointly enforces the boundary of predicted semantic regions to be aligned with that of the ground truth regions.

Lastly, we leverage the eigenvectors of Laplacian of affinity matrices that is constructed by features learned in a self-supervised manner, to decompose inputs into eigensegments.
They are then fused with learning-based predictions from the trained model.
The fusion outputs are taken as the final predictions of the framework.

As illustrated in Fig.~\ref{fig:teaser}, compared with \cite{li2022language}, the predictions of our approach are better aligned with the shapes of objects.

We also demonstrate the effectiveness of our approach with elaborate experiments on PASCAL-${5}^i$ and COCO-${20}^i$, the results of which show that our method outperforms former state-of-the-arts \cite{li2022language,bucher2019zero,xian2019semantic,wang2020few,min2021hypercorrelation,nguyen2019feature} by large margins.
By examining a) the correlation between \textbf{shape compactness} of target object and IoU and b) the correlation between the \textbf{language embedding locality} and IoU, we discover the large impacts on the performance brought by the distribution of language anchors and object shapes.
Via extensive analyses, we demonstrate the effectiveness and generalization of SAZS framework's shape perception for segmenting semantic categories in the open world.



\begin{figure*}[h]
    \centerline{
    \includegraphics[width=0.89\linewidth]
    {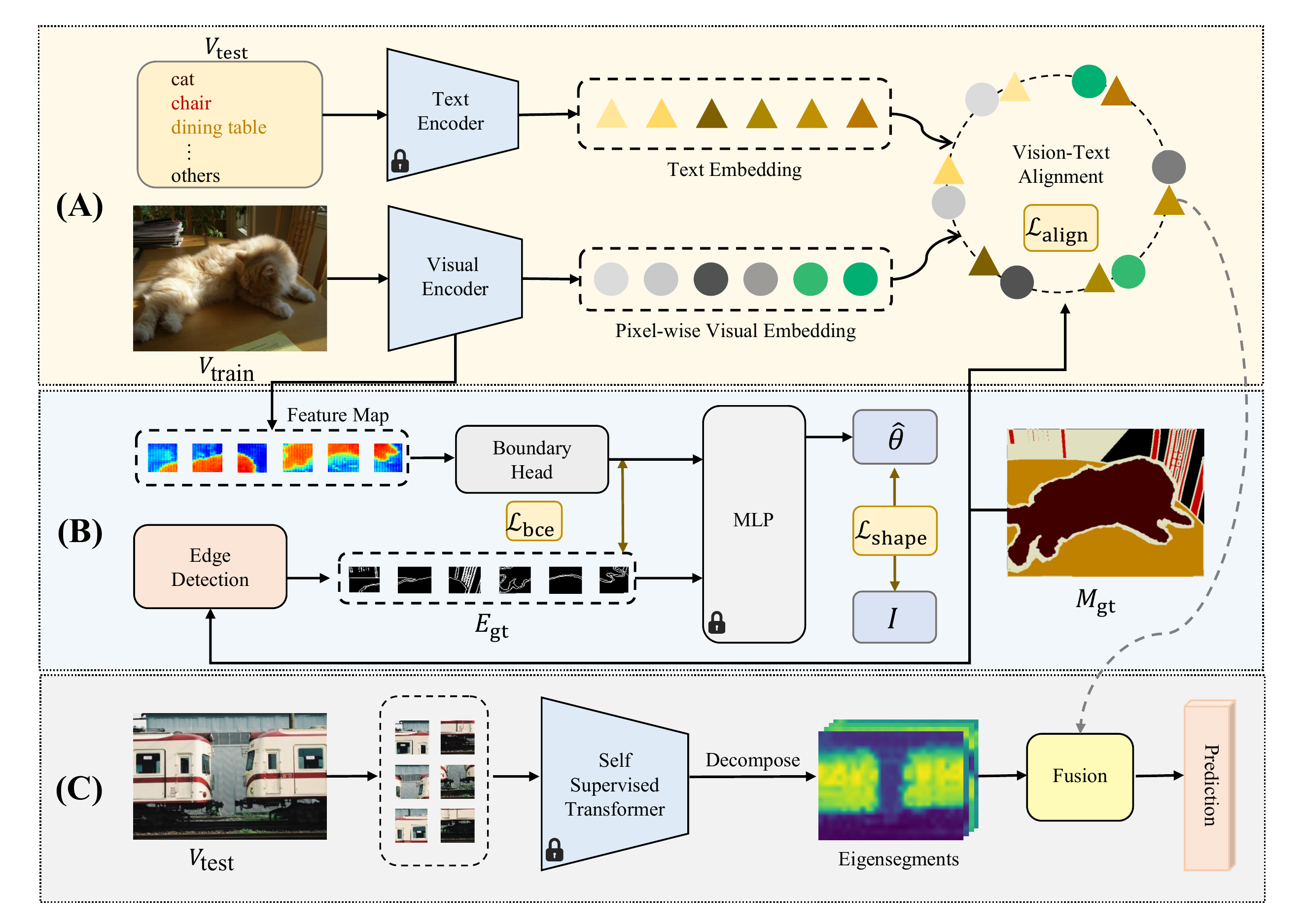}}
    \caption{The overview of SAZS framework. SAZS addresses the task of zero-shot semantic segmentation, which aims to segment the test set image $V_{\text{test}}$ by open-set categories without additional training of the network. During training, \textbf{(A)} the input image $V_{\text{train}}$ is transformed into pixel-wise visual embeddings which are aligned with the text embeddings of training categories $T_{\text{train}}$, according to the ground-truth semantic maps $M_{\text{gt}}$. The text embeddings are obtained by the pre-trained text encoder of CLIP \cite{radford2021learning} and serve as optimization anchors of the CLIP feature space. \textbf{(B)} In order to aggregate shape priors contained in the input image, SAZS jointly trains on the constraint task of boundary detection by comparing the ground-truth boundaries and the predictions of boundary heads of the visual encoder. \textbf{(C)} During inference, in order to reduce the domain gap between seen and unseen categories, SAZS fuses the pixel-wise predictions of the neural networks with eigensegments obtained by non-learning-based spectral analysis. Note that, modules marked with the lock icon are pre-trained and not optimized during training of SAZS.}
    \label{fig:main}
\end{figure*}


\section{Related works}

\subsection{Zero-Shot Semantic Segmentation}

The main goal of the zero-shot semantic segmentation task(ZSS) is to
perform pixel-wise predictions for objects that are unseen during training.
Recent works on ZSS have seen two main branches: the generative methods and the discriminative methods.

The generative methods 
\cite{bucher2019zero,gu2020context,li2020consistent} produce  synthesized features for unseen categories.

ZS3Net\cite{bucher2019zero} utilizes a generative model to create a visual representation of objects that were not present in the training data. This is achieved by leveraging pre-trained word embeddings.
CaGNet\cite{gu2020context} highlights the impact of contextual information on pixel-level features through a network learning to generate specific contextual pixel-level features.
In CSRI\cite{li2020consistent}, constraints are introduced to the generation of unseen visual features by exploiting the structural relationships between seen and unseen categories. 

As for the discriminative methods, SPNet\cite{xian2019semantic} leverages similarities between known categories to transfer learned representations to other unknown categories.
Baek et al. \cite{baek2021exploiting} employ visual and semantic encoders to learn a joint embedding space with the semantic encoder converting the semantic features into semantic prototypes.
Naoki et al.\cite{kato2019zero} introduce variational mapping by projecting the class embeddings from the semantic to the visual space.
Lv et al. \cite{lv2020learning} present a transductive approach using target images to mitigate the prediction bias towards seen categories.
LSeg\cite{li2022language} proposes a language-driven ZSS model, mapping pixels and names of labels into a shared embedding space for pixel-wise dense prediction. 

Though much pioneering efforts have been spent, the dense prediction task requires fine shape delineation while most existing vision-language models are trained with image-level language descriptions.
How to effectively address these problems is the focus of our work.
\subsection{Shape-aware Segmentation}
Shape awareness is beneficial to dense prediction tasks. Most of the semantic segmentation methods\cite{long2015fully,ronneberger2015u,chen2017deeplab,wang2020deep} cannot preserve object shapes since they only focus on feature discriminativeness but ignore proximity between central and other positions.

Meanwhile, SGSNet\cite{zhang2020semi} takes a hierarchical approach to aggregating the global context when modeling long-range dependencies, considering feature similarity and proximity to preserve object shapes. ShapeMask\cite{kuo2019shapemask} refines the coarse shapes into instance-level masks. The shape priors provide powerful clues for prediction.
Gated-SCNN\cite{takikawa2019gated} proposes a two-stream architecture for semantic segmentation that explicitly captures shape information as a separate processing branch. The key point is to enable the interactive flow of information between the two networks, allowing the shape stream to focus on learning and processing of edge information. Liu et al. \cite{liu2017learning} construct the spatial propagation networks for learning the affinity matrix. The affinity matrix allows a tractable modeling of the dense, global pairwise relationships of pixels.

\subsection{Spectral Methods for Segmentation}
Among different segmentation schemes, spectral methods for clustering   employ the  eigenvectors of a matrix derived from the distance between points, which have been successfully used in many applications.

Shi et al. \cite{shi2000normalized} regard image segmentation as a  graph partitioning problem, which proposes a novel global criterion, the normalized cut, to segment the image. Soft segmentations\cite{aksoy2018semantic} are generated automatically by fusing high-level and low-level image features in a graph structure. The purpose of constructing this graph is to enable the corresponding Laplacian matrix and its eigenvectors to reveal semantic objects and soft transitions between objects.
In our work, we utilize the eigensegments obtained by self-supervised spectral decomposition with the network outputs as the framework's predictions to avoid the bias of the learning-based model on the training set and further improve shape-awareness.

\subsection{Vision-Language Modeling}

An extensive group of works have investigated the zero-shot semantic segmentation task. The key idea behind many of these works is to exploit priors encoded in pretrained word embeddings to generalize to unseen classes and achieve dense predictions \cite{zhao2017open,bucher2019zero,gu2020context,xian2019semantic,kato2019zero,hu2020uncertainty,li2020consistent,pastore2021closer,li2022language,li2022toist,li2022understanding,jin2023adapt,gao2021rcd}, such as word2vec\cite{miller1995wordnet}, GloVe\cite{pennington2014glove} or BERT\cite{devlin2018bert}. CLIP\cite{radford2021learning} has recently demonstrated impressive zero-shot generalization in various image-level classification tasks. As a result, several works have since exploited the vision-language embedding space learned by CLIP\cite{radford2021learning} to enhance dense prediction capabilities \cite{rao2022denseclip,li2022language,xu2021simple}. CLIP develops contrastive learning with a large-capacity language and visual feature encoder to train extremely robust models for zero-shot image classification. But the performance of large-scale pre-trained vision encoders transferred to pixel-level classification work is unsatisfactory. Unfortunately, the direct utilization of the extracted image-level vision-language features ignores the discrepancy between image-level and the pixle-level dense predicition task, the latter of which is the focus of our work. According to our study, the shape-aware prior and supervision can bridge this discrepancy and get more accurate segmentation results.


\section{Methods}
\label{sec:formatting}

The goal of zero-shot semantic segmentation is to extend semantic segmentation task to the unseen categories other than those in the training datasets.
One potential approach to introduce extra priors is to leverage pre-trained vision-language models, yet most of these models focus on the image-level prediction and fail to transfer to dense prediction tasks.

To this end, we propose a novel method named \textbf{S}hape-\textbf{A}ware \textbf{Z}ero-\textbf{S}hot Semantic Segmentation (\textbf{SAZS}).

This approach leverages the rich language priors contained in the pre-trained CLIP \cite{radford2021learning} model, while also exploiting the proximity between local regions to perform the boundary detection task with constraints. Meanwhile, we utilize spectral decomposition of self-supervised visual features to improve our approach's sensitivity to shape, and integrate this with pixel-wise prediction.

The overall pipeline of our methods is depicted in Fig. \ref{fig:main}. The input image is first transformed by an image encoder into pixel-wise embeddings, which are then aligned with precomputed text embeddings obtained by the text encoder of pre-trained CLIP model (Part A in Fig. \ref{fig:main}).
Meanwhile, an extra head in the image encoder is used to predict the boundaries in patches, which are optimized towards the ground-truth edges obtained from segmentation ground truths (Part B in Fig. \ref{fig:main}).
In addition, we further exploit proximity of local regions during inference by decomposing the image by spectral analysis and fusing the output eigensegments with class-agnostic segmentation results (Part C in Fig. \ref{fig:main}).

In the following section, we first formally define the addressed task and introduce the notations in Sec. \ref{subsec:problem}. Then we describe the loss designs for the vision-language alignment and boundary prediction in Sec. \ref{subsec: vt align} and Sec. \ref{subsec: shape opt}, respectively. The inference pipeline involving spectral decomposition of the proposed affinity matrix is introduced in Sec. \ref{subsec: infer}. 

\subsection{Task Deﬁnition}
\label{subsec:problem}

Following HSNet\cite{min2021hypercorrelation}, we denote the training set by $\mathcal{D}_{\text {train }}=\left\{({I}, {M}, \mathcal{S})\right\}$ and testing set by $\mathcal{D}_{\text {test }}=\left\{({I}, {M}, \mathcal{U})\right\}$, where ${I} \in \mathbb{R}^{H \times W \times 3}$ and ${M} \in \mathbb{R}^{H \times W \times C}$ denote an input image and the corresponding ground-truth semantic mask with digit encoding.
$\mathcal{S}$ denotes the set of $K$ potential labels in ${I}$, while $\mathcal{U}$ denotes the set of unseen categories during testing. The two sets are strictly exclusive in our setting (i.e., $\mathcal{S} \cap \mathcal{U}=\emptyset$).

Before inferencing on $\mathcal{D}_{\text {test}}$ targeting $\mathcal{U}$, the model is trained on $\mathcal{D}_{\text {train}}$ with ground-truth labels from $\mathcal{S}$.
This means the categories in the test set are never seen during training, making the task formulated in a zero-shot setting. 
Once the model is well-trained, it is expected to generalize to unseen categories and achieve high performance for dense prediction of target objects in the open world.


\subsection{Pixel-wise Vision-Language Alignment}
\label{subsec: vt align}
Comparing distances between pixel features and different text anchor features in the shared feature space is a straightforward approach for zero-shot semantic segmentation.
However, while the pioneer work CLIP \cite{radford2021learning} introduces a shared feature space for visual and text inputs, the image-level CLIP visual encoder is infeasible for dense prediction tasks since fine details in images, as well as the correlation between pixels, are lost.
In this section, we describe our approach to address this issue by optimizing a dense visual encoder separate of CLIP and enforcing the pixel-wise output features towards the text anchors in the CLIP feature space during training.

\subparagraph{Visual Encoder}

We employ  Dilated residual networks (DRN)\cite{yu2017dilated} and Dense Prediction Transformers (DPT) \cite{ranftl2021vision} to encoder images into pixel-level embeddings.
More specifically, an input image of size $H \times W \times 3$ is first processed with standard augmentation to  $\tilde{H} \times \tilde{W}\times 3$ and then passed as input to the visual encoder, resulting in a feature map $\mathcal{F}_{V}\in \mathbb{R}^{\tilde{H} \times \tilde{W} \times D}$, where $D$ is the feature size in DPT.

\subparagraph{Text Encoder}

While most concurrent methods use digit labels (e.g., 0, 1, 2) to represent categories, we take embeddings of the category names (e.g. "airplane", "cat" ) as the anchors of  feature space. 
These embeddings are obtained with the CLIP text encoder.
Specifically, we adopt the pre-trained CLIP text encoder to map the names of $K$ categories from $\mathcal{S}$ into CLIP feature space as the anchor features $\mathcal{F}_{T} \in \mathbb{R}^{K\times D}$, which is later used as targets for optimization.

Note that, the visual features $\mathcal{F}_{V}$ and the text features $\mathcal{F}_{T}$ have the same dimension $D$.

\subparagraph{Vision-Language Alignment}
To enforce vision-language alignment, the distances between pixels and corresponding semantic category should be minimized while the distances between pixels and other categories should be maximized.
Under the assumption that pixel-wise vision and language features are embedded in the same feature space, we leverage the cosine similarity $\left\langle\cdot,\cdot\right\rangle$ as the quantitative distance metric between features and propose the alignment loss as the sum of cross entropy losses over seen classes of all pixels:

\begin{equation}
\label{eq:lseg}
\mathcal{L}_{\rm align}=
\sum_{i, j}^{\tilde{H}, \tilde{W}}
\left(-\log \frac{e^{\left\langle \mathcal{F}_{V}\left[i, j\right], \mathcal{F}_{T}\left[k_{ij}\right]\right\rangle}}{\sum_{k^{\prime}=1}^{|S|} e^{\left\langle \mathcal{F}_{V}\left[i, j\right], \mathcal{F}_{T}\left[k^\prime\right]\right\rangle}}\right)
\end{equation}
In Eq. \ref{eq:lseg}, $\mathcal{F}_{V}\left[i, j\right]$ denotes pixel visual feature at position $(i,j)$, $\mathcal{F}_{T}\left[k\right]$ denotes $k$-th text anchor features and $k_{ij}$ denotes index of ground-truth category of pixel at $(i, j)$.

\subsection{Shape Constraint}
\label{subsec: shape opt}

Since CLIP is trained on an image-level task, simply leveraging the priors in the CLIP feature space may be insufficient for dense prediction tasks.
To address this issue, we introduce boundary detection as a constraint task, so that the visual encoder is able to aggregate finer information contained in images.
Inspired by InverseForm \cite{borse2021inverseform}, we address this constraint task by optimizing the affine transformation between ground-truth edges and edges in feature maps towards identity transformation matrix.

More specifically, as shown in Fig. \ref{fig:main}, we extract middle-layer features of the visual encoders and split them into patches.
On the one hand, the ground truth edges within the patches are obtained by applying Sobel operator on ground truth semantic masks.
On the other hand, the feature patches are processed by a boundary head.
Then, we calculate the affine transform matrix $\hat{\theta}_i$ for the $i$-th patch between ground-truth edges and processed feature patches with a pre-trained MLP.
Note that, this MLP is trained in advance with edge masks and not optimized during our method's training.
We optimize this affine transform matrix towards identity matrix by:
\begin{equation}
\mathcal{L}_{\rm shape}=\frac{1}{T} \sum_{i=1}^T\left|\hat{\theta}_i-I\right|_F
\end{equation}
where $T$ denotes the number of patches and $\left|\cdot\right|$ denotes Frobenius norm.

Furthermore, we directly calculate the binary cross entropy loss $\mathcal{L}_{\rm {bce}}$ between the predicted edge masks of the whole image and corresponding ground truths to further optimize the performance of boundary detection.

After jointly training on the task of boundary detection, the visual encoder is enabled to collect and leverage shape priors in the input images. Ablation studies detailed later show that shape awareness introduced by $\mathcal{L}_{\rm {shape}}$ and $\mathcal{L}_{\rm bce}$ brings about notable improvements.

Finally, the overall loss to optimize during training is:
\begin{equation}
\mathcal{L}=\mathcal{L}_{\rm{align}}+\lambda_1 \mathcal{L}_{\rm {shape }}+\lambda_2 \mathcal{L}_{\rm {bce }}
\end{equation}
where $\lambda_1$ and $\lambda_2$ are loss weights.

\subsection{Self-supervised Spectral Decomposition}
\label{subsec: infer}

We seek to decompose the input images into eigensegments with clear boundaries in an unsupervised manner, and then fuse these eigensegments with the predictions of the neural networks in the fusion module in Fig. \ref{fig:main} .

The derivation of affinity matrix is the key to spectral decomposition. Following Melas-Kyriazi et al. \cite{melas2022deep}, we first leverage the features $f$ from the attention block of the last layer of a pre-trained self-supervised transformer (i.e., DINO \cite{caron2021emerging}). The affinity between pixel $i$ and $j$ is defined as:
\begin{equation}
Z_{\rm {sem }}(i, j)=f_i\cdot f_j^T
\end{equation}
Note that, the self-supervised transformer is only used during inference and its weights are not optimized.

While the affinities derived from transformer features are rich in semantic information, the low-level proximity including color similarity and spatial distance is missing.
Inspired by image matting \cite{chen2013knn,levin2007closed}, we first transfrom the input image into the HSV color space: $X(i)=(\cos (h), \sin (h), s, v, x, y)_i$, where $h, s, v $ are the respective HSV coordinates and $(x, y)$ are the spatial coordinates of pixel $i$.
Then, the affinity between pixels is defined as

\begin{equation}
Z_{\rm{shape}}(i, j)=1-\|X(i)-X(j)\|_2, \ j \in \mathrm{KNN}{(i)}
\end{equation}
where $\|\cdot\|_2$ denotes 2-norm.
The overall affinity matrix is defined as the weighted sum of the two:
\begin{equation}
Z(i, j)= Z_{\rm{sem}} + \lambda\cdot Z_{\rm{shape}}
\end{equation}

With the affinity matrix, we now can compute the eigenvectors of the Laplacian $L$ of the affinity matrix, which are used to decompose the image into multiple eigensegments.
\subsection{Inference}

Given an image for inference, we first encode the phrases of the categories using the pre-trained text encoder CLIP and obtain textual features $\mathcal{F}_{\mathrm T} \in \mathbb{R}^{C\times D}$ for ${C}$ categories, each of which is represented by a $D$-dimension embedding. Then we leverage the trained visual encoder to obtain the visual feature map $\mathcal{F}_{\mathrm V}\in \mathbb{R}^{\tilde{H} \times \tilde{W} \times D}$. The final logits $\hat{F_{ij}}=\mathcal{F}_{\mathrm V}(i,j) \cdot \mathcal{F}_{\mathrm T}^T $ are calculated as the cosine similarities between the visual feature map and textual features. In the mean time, we employ the pre-trained DINO to extract semantic features in an unsupervised manner and calculate the top $K$ spectral eigensegments ${E_k}$ ($K=5$ in our implementation).
The final prediction results are generated by the fusion module, which selects from the sets of predictions according to the maximal IoU (denoted as $\Phi_{\text {FUSE}}$) of the ${E_k}$ and $\operatorname{argmax} \hat{F_{i j}}$.

\begin{equation}
\operatorname{Pred}_{i j}=\Phi_{\text {FUSE }}\left(E_k, \operatorname{argmax} \hat{F}_{i j}\right)\, \quad k \in\{0,1, \cdots K\}
\end{equation}

\section{Experiments}
\subsection{Datasets}
We extensively evaluate our method on two datasets dedicated for the task of zero-shot semantic segmentation: PASCAL-$5^i$ \cite{everingham2015pascal} and COCO-${20}^i$ \cite{lin2014microsoft}.
    Built upon PASCAL VOC 2012 \cite{everingham2015pascal} and augmented by SBD \cite{hariharan2011semantic}, PASCAL-$5^i$ contains 20 categories which are further divided into 4 folds denoted by $5^0$, $5^1$, $5^2$ and $5^3$. Each image is annotated with 5 categories within each fold.
Similarly, based on MS COCO \cite{lin2014microsoft}, COCO-${20}^i$ is a more challenging dataset with 80 categories divided into four folds denoted by $20^0$, $20^1$, $20^2$ and $20^3$, and each of the four folds contains 20 categories.
Of the four folds in the two datasets, one is used for evaluation (i.e., the target fold) while the other three are used for training.
In the following section, we denote each experiment setup by the target fold. 

Following prior literature on zero-shot semantic segmentation, we adopt mean intersection over union (mIoU) and foreground-background IoU (FBIoU) as the evaluation metrics. 
Specifically, mIoU is the average of IoUs of the categories in target fold and FBIoU is the average of foreground IoU and background IoU.

 \begin{table*}
    \scriptsize
	\centering
            \begin{tabular}{ccc|cccccc|cccccc}
            \toprule
            \multirow{2}{*}{\textbf{Method}}& \multirow{2}{*}{\textbf{Backbone}} & \multirow{2}{*}{\textbf{Setting}} & \multicolumn{6}{c}{\textbf{PASCAL-${5}^i$}}& \multicolumn{6}{|c}{\textbf{COCO-${20}^i$}} \\
             &  &  & $5^{0}$ & $5^{1}$ & $5^{2}$ & $5^{3}$ & mIoU &FBIoU &$20^{0}$ & $20^{1}$ & $20^{2}$ & $20^{3}$ & mIoU &FBIoU\\
            \midrule FWB\cite{nguyen2019feature} & ResNet& 1-shot & $51.3$ & $64.5$ & $56.7$ & $52.2$ & $56.2$ & $-$ &  $17.0$ & $18.0$ & $21.0$ & $28.9$ & $21.2$&$-$\\
            DAN\cite{wang2020few} & ResNet & 1-shot & $54.7$ & $68.6$ & $57.8$ & $51.6$ & $58.2$ & $71.9$ &  $-$ & $-$ & $-$ & $-$ & $24.4$&$62.3$\\
            PFENet\cite{tian2020prior} &ResNet & 1-shot & $60.5$ & $69.4$ & $54.4$ & $55.9$ & $60.1$ & $72.9$& $36.8$ & $41.8$ & $38.7$ & $36.7$ & $38.5$&$63.0$\\
            HSNet\cite{min2021hypercorrelation} &ResNet & 1-shot & $67.3$ & $72.3$ & $62.0$ & $63.1$ & $66.2$ &$77.6$  & $37.2$ & $44.1$ & $42.4$ & $41.3$ & $41.2$&$69.1$\\
            \midrule SPNet\cite{xian2019semantic} &ResNet & zero-shot & $23.8$ & $17.0$ & $14.1$ & $18.3$ & $18.3$ &$44.3$ & $-$ & $-$ & $-$ & $-$ & $-$&$-$\\
            ZS3Net\cite{bucher2019zero} & ResNet & zero-shot & $40.8$ & $39.4$ & $39.3$ & $33.6$ & $38.3$  &$57.7$ & $18.8$ & $20.1$ & $24.8$ & $20.5$ & $21.1$ &$55.1$ \\
            LSeg\cite{li2022language}&ResNet  & zero-shot & $52.8$ & $53.8$ & $44.4$ & $38.5$ & $47.4$ &$64.1$ & $22.1$ & $25.1$ & $24.9$ & $21.6$ & $23.4$& $57.9$\\
            \textbf{Ours} & DRN & zero-shot & $\mathbf{57.3}$ & $\mathbf{60.3}$ & $\mathbf{58.4}$ & $\mathbf{45.9}$ & $\mathbf{55.5}$ & $\mathbf{66.4}$ & $\mathbf{34.2}$ & $\mathbf{36.5}$ & $\mathbf{34.6}$ & $\mathbf{35.6}$ & $\mathbf{35.2}$& $\mathbf{58.4}$\\
            
            \midrule LSeg\cite{li2022language} & ViT-L & zero-shot & $61.3$ & $63.6$ & $43.1$ & $41.0$ & $52.3$  &$67.6$ & $28.1$ & $27.5$ & $30.0$ & $23.2$& $27.2$& $\mathbf{59.9}$\\
            \textbf{Ours} & ViT-L & zero-shot & $\mathbf{62.7}$ & $\mathbf{64.3}$ & $\mathbf{60.6}$ & $\mathbf{50.2}$ & $\mathbf{59.4}$ & $\mathbf{69.0}$  & $\mathbf{33.8}$ & $\mathbf{38.1}$ & $\mathbf{34.4}$ & $\mathbf{35.0}$ & $\mathbf{35.3}$ & $58.2$\\
            \bottomrule
            \end{tabular}
\vspace{-4pt}
\caption{The performances of SAZS and baselines evaluated on PASCAL-${5}^i$ and COCO-${20}^{i}$}
\vspace{-12pt}
\label{tab:all_result}
\end{table*}

\subsection{Implementation Details}
In our experiments, we employ the pre-trained CLIP-ViT-B/32 as the text encoder.
Background or unknown category is regarded as "others" when mapped from text to CLIP features.
The visual encoder is implemented by DRN \cite{yu2017dilated} or DPT\cite{ranftl2021vision} with ViT \cite{dosovitskiy2020image} as the backbone.
When training on the task of boundary detection, each feature map for the shape boundary and the corresponding ground truth are splitted into $3 \times 6$ patches. Each patch pair is then fed into the MLP in Part B of Fig.~\ref{fig:main} to calculate the affine transformation matrix.

During training, the network is optimized by an SGD optimizer with a momentum of 0.95 and a learning rate of $5\times10^{-5}$ decayed by a polynomial scheduler.
With ViT as the backbone of visual encoder, the training process finishes within 5 epochs on 4 NVIDIA Tesla V100 GPUs with a batch size of 6.

\subsection{Results}

The proposed method SAZS has been evaluated on the PASCAL-$5^i$ and COCO-${20}^i$ datasets under zero-shot settings, alongside several baselines for comparison.
The performances are reported in Tab.~\ref{tab:all_result}.
With DRN as the visual encoder backbone, our method achieves large margins over the strong baseline LSeg \cite{li2022language}, with mIoU improved by $6.1 \%$ and $4.8 \%$ on PASCAL-$5^i$ and COCO-${20}^i$ respectively.
Our model also outperforms LSeg \cite{li2022language} by large margins with the ViT backbone underlying DPT, with mIoU improved by $7.2 \%$ and $11.2\%$. 
The performance enhancements of SAZS remain consistent across different visual encoder choices, highlighting its effectiveness.

\begin{figure}[h]
    \centerline{
    \includegraphics[width=1.0\linewidth]
    {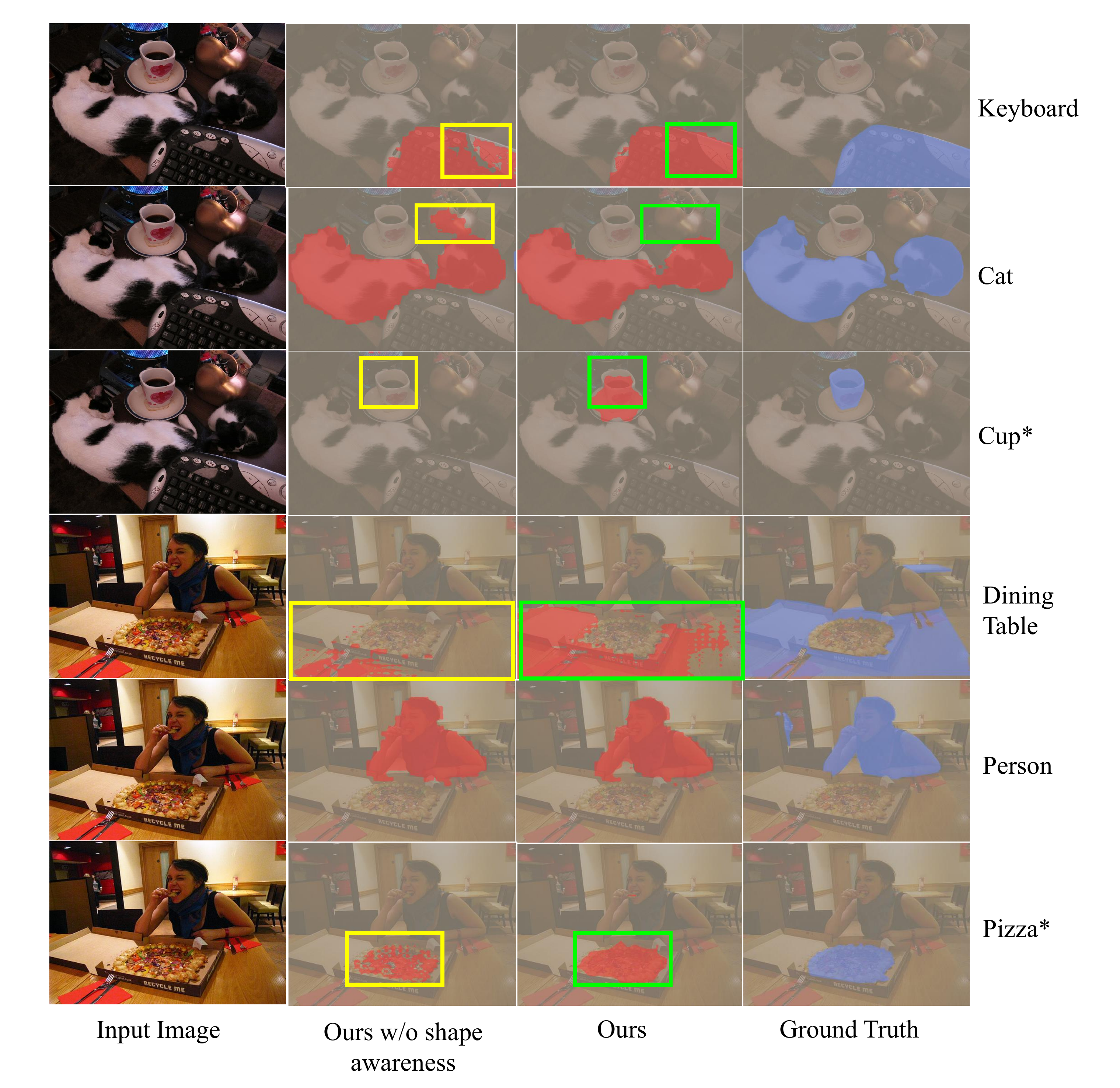}}
    \caption{Qualitative results of COCO-${20}^i$. The first and last columns are the input images and the corresponding ground-truth semantic masks for different categories. The second and the third columns are the predictions by SAZS without and with shape awareness, respectively. \textbf{*} denotes unseen categories during training phase) and yellow boxes mark poorly segmented regions. }
    \label{fig:coco_qual}
\end{figure}

\begin{figure}[h]
    \centerline{
    \includegraphics[width=1.0\linewidth,height=0.7\linewidth]
    {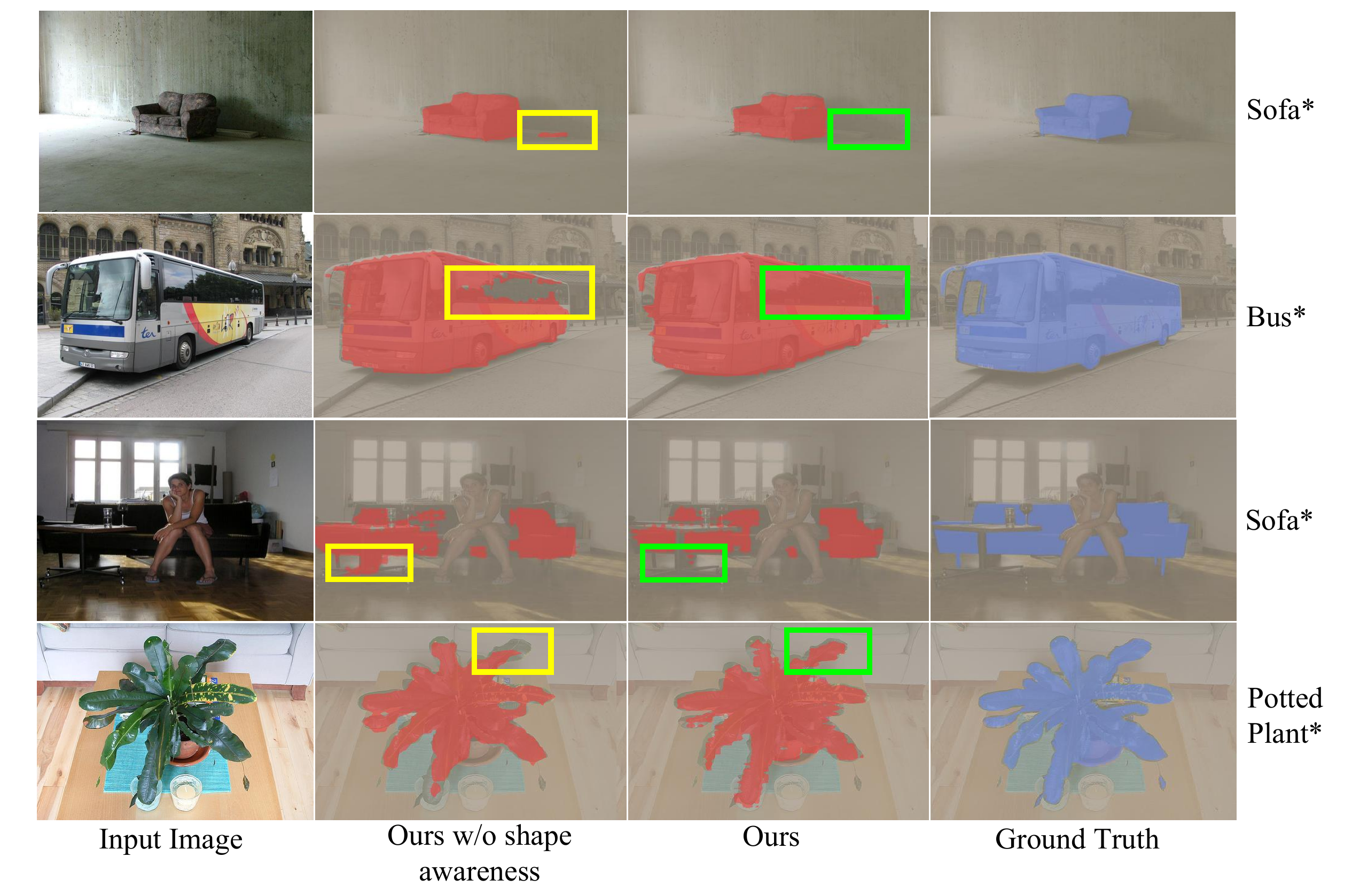}}
    \caption{Qualitative comparison results of PASCAL-${5}^i$. The first and last columns are the input images and the corresponding ground-truth semantic masks for different categories. The second and the third columns are the predictions by SAZS without and with shape awareness, respectively. \textbf{*} denotes categories categories unseen during training phase) and yellow boxes mark poorly segmented regions.}
    \label{fig:pascal_qual}
\end{figure}

 In addition, we conduct cross-dataset validation by training on the COCO-${20}^i$ dataset and testing on  PASCAL-${5}^i$. As shown in Table \ref{tab:cross}, our method outperforms OpenSeg\cite{ghiasi2021open} and LSeg+\cite{li2022language} in zero-shot dense prediction tasks with clear margins. It is worth noting that all three methods are trained on a larger semantic segmentation dataset (COCO-${20}^i$). These performance gaps demonstrate the generalization ability of our shape-aware training framework across datasets.
 \begin{table}
    \scriptsize
	\centering
	\begin{tabular}{c|c|c|c|c}
		\toprule
		\textbf{Model} & \textbf{Backbone} & \textbf{external dataset} & \ \textbf{target dataset} & \textbf{PASCAL-${5}^i$} \\
		\midrule
            LSeg   & ViT-L & \XSolidBrush  & \Checkmark (seen classes) &  52.3\\
        \midrule
        SPNet   & ResNet &  \XSolidBrush  & \Checkmark (seen classes)  & 18.3\\
            ZS3Net  & ResNet & \XSolidBrush  & \Checkmark (seen classes) & 38.3\\
            LSeg  & ResNet &  \XSolidBrush  & \Checkmark (seen classes)&  47.4\\
            \midrule
            $$\text { LSeg+ }$$   & ResNet &  COCO  & \XSolidBrush  &  59.0\\
            OpenSeg\cite{ghiasi2021open}   & ResNet & COCO  &\XSolidBrush   & 60.0\\
            Ours   & DRN &   COCO &  \XSolidBrush  & \textbf{62.7}  \\
		\bottomrule
	\end{tabular}
	\vspace{-4pt}
	\caption{The cross dataset mIoU results of our model and previous SOTA methods on PASCAL-${5}^i$.}
	\vspace{-12pt}
	\label{tab:cross}
\end{table}
We also provide qualitative results for the proposed method SAZS.
in Fig. \ref{fig:coco_qual} and Fig. \ref{fig:pascal_qual}.
In these figures, we illustrate the predictions of SAZS with and without shape awareness on COCO-${20}^i$ and PASCAL-${5}^i$ respectively, showing its ability to make precise predictions on both seen and unseen categories.

\subsection{Ablation Study}
To further demonstrate the effectiveness of design choices in our approach, we perform detailed ablation studies by evaluating our method with or without shape constraint during training as well as the fusion of network predictions with eigensegments.
Results on PASCAL-${5}^i$ are reported in Tab.~\ref{tab:abl_p} and results on COCO-${20}^i$ are reported in Tab.~\ref{tab:abl_coco_vit} and Tab.~\ref{tab:abl_coco_drn}.

\paragraph{Effects of Shape-awareness}
The motivation for auxiliary constraint $\mathcal{L}_{\rm shape}$ is to learn the shape priors of images contained in the target boundaries.
We observe that without training on the constraint task of boundary detection, the performances of the proposed method tend to decline.
Specifically as reported in Tab.~\ref{tab:abl_coco_vit} and Tab.~\ref{tab:abl_coco_drn}, the mIoU of SAZS drops by $1.4 \%$ and by $1.5 \%$ with ViT and DRN backbone on COCO-${20}^i$ when training without $\mathcal{L}_{\rm shape}$.
The performance gaps clearly indicate the significant role played by shape-awareness in the proposed SAZS framework.

\paragraph{Effect of Fusion with Spectral Eigensegments}

We also demonstrate  the importance of fusing with spectral eigensegments during inference.
Without the fusion module, the mIoU dramatically decreases by $7.0\%$ on PASCAL-${5}^i$ and by $6.2\%$ (ViT backbone) and $8.6\%$ (DRN backbone) on COCO-${20}^i$, as reported in Tab.~\ref{tab:abl_p}, Tab.~\ref{tab:abl_coco_vit} and Tab.~\ref{tab:abl_coco_drn}.
These large margins indicate that eigensegments obtained by spectral decomposition of the affinity matrices largely suppress the bias on the training dataset and seen categories.

\begin{table}
    \scriptsize
	\centering
            \begin{tabular}{c|cc|ccccc}
            \toprule
            \textbf{Model} &\textbf{ Fusion} & $ \mathcal{L}_{\rm {shape }}$ & $5^{0}$ & $5^{1}$ & $5^{2}$ & $5^{3}$ & $\textbf{mIoU}$ \\
             \midrule
             $\text{SAZS}$ &\Checkmark & \Checkmark & $62.7$ & $\mathbf{64.3}$ & $\mathbf{60.6}$ & $\mathbf{50.2}$ & $\mathbf{59.4}$ \\
             $\text{SAZS}$ &\Checkmark &  & $\mathbf{63.1}$ & $62.4$ & $59.0$ & $49.2$ & $58.4$\\
              $\text{SAZS}$ && \Checkmark & $59.7$ & $63.4$ & $44.3$ & $42.2$ & $52.4$ \\
              $\text{SAZS}$ & & & $59.2$ & $61.9$ & $43.8$ & $41.9$ & $51.7$ \\
               \midrule
               LSeg\cite{li2022language} & & &$61.3$ & $63.6$ & $43.1$ & $41.0$ & $52.3$  \\
            \bottomrule
            \end{tabular}
\vspace{-4pt}
\caption{Ablation study on PASCAL-${5}^i$ (ViT backbone)}
\vspace{-12pt}
\label{tab:abl_p}%
\end{table}%

\begin{table}
    \scriptsize
	\centering
            \begin{tabular}{c|cc|ccccc}
             \toprule
             $\textbf{Model}$ &\textbf{Fusion} & $ \mathcal{L}_{\rm {shape }}$ & $20^{0}$ & $20^{1}$ & $20^{2}$ & $20^{3}$ & $\textbf{mIoU}$\\
            \midrule
            $\text{SAZS}$ &\Checkmark & \Checkmark & $\mathbf{33.8}$ & $38.1$ & $\mathbf{34.4}$ & $\mathbf{35.0}$ & $\mathbf{35.3}$\\
             $\text{SAZS}$ &\Checkmark &  & $33.3$ & $\mathbf{39.0}$ & $33.9$ & $32.7$ & $34.7$\\
             $\text{SAZS}$ & & \Checkmark& $30.0$ & $30.4$ & $27.5$ & $28.5$ & $29.1$  \\
              $\text{SAZS}$ & &  & $26.3$ & $32.0$ & $26.2$ & $26.2$ & $27.7$ \\
              \midrule
              LSeg\cite{li2022language} & &&  $28.1$ & $27.5$ & $30.0$ & $23.2$& $27.2$\\
            \bottomrule
            \end{tabular}
\vspace{-4pt}
\caption{Ablation study on COCO-${20}^i$ (ViT backbone)}
\vspace{-12pt}
\label{tab:abl_coco_vit}
\end{table}
\begin{table}
    \scriptsize
	\centering
            \begin{tabular}{c|cc|ccccc}
            \toprule
            $\textbf{Model}$&\textbf{Fusion} & $ \mathcal{L}_{\rm {shape }}$ & $20^{0}$ & $20^{1}$ & $20^{2}$ & $20^{3}$ & $\textbf{mIoU}$\\
            \midrule
            $\text{SAZS}$&\Checkmark & \Checkmark & $\mathbf{34.2}$ & $36.5$ & $\mathbf{34.6}$ & $\mathbf{35.6}$ & $\mathbf{35.2}$ \\
             $\text{SAZS}$&\Checkmark &  & $33.7$ & $\mathbf{38.2}$ & $33.4$ & $35.5$ & $35.2$ \\
            $\text{SAZS}$& & \Checkmark & $28.4$ & $27.6$ & $25.4$ & $25.1$ & $26.6$  \\
              $\text{SAZS}$&&  & $24.2$ & $28.5$ & $24.4$ & $23.3$ & $25.1$ \\
            \midrule
            LSeg\cite{li2022language} & && $22.1$ & $25.1$ & $24.9$ & $21.6$ & $23.4$ \\
            \bottomrule
            \end{tabular}
\vspace{-4pt}
\caption{Ablation study on COCO-${20}^i$ (DRN backbone)}
\vspace{-12pt}
\label{tab:abl_coco_drn}
\end{table}

\subsection{Ablation of $Z_{\rm{sem}}$ and $Z_{\rm{shape}}$}
We conduct an ablation experiment on the PASCAL-${5}^i$ dataset to investigate the effects of $Z_{\rm{sem}}$ and $Z_{\rm{shape}}$ in our fusion module. As shown in Table \ref{tab:lambda}, both $Z_{\rm{sem}}$ and $Z_{\rm{shape}}$ contribute to improved segmentation performance, but using $Z_{\rm{sem}}$ alone yields better results than using $Z_{\rm{shape}}$ alone. While the segmentation performance obtained by combining the two is slightly higher than that achieved with $Z_{\rm{sem}}$ alone, using both requires fine-tuning the hyper-parameter $\lambda$, which can be unstable and requires additional effort. 
\begin{table}
    \scriptsize
	\centering
	\begin{tabular}{c|c|cc|c}
		\toprule
		\textbf{Model} & \textbf{external dataset}&\textbf{ $Z_{\rm{shape}}$} & \textbf{ $Z_{\rm{sem}}$}  & \textbf{PASCAL-${5}^i$} \\
		\midrule
              SAZS& COCO & &   &  58.4\\
             SAZS &COCO & \Checkmark  &    &  58.6\\
		    SAZS&COCO&  & \Checkmark  &  62.7\\
		\bottomrule
	\end{tabular}
	\vspace{-4pt}
	\caption{Impact of $Z_{\rm{shape}}$ and  $Z_{\rm{sem}}$ Of fusion module (in the cross-dataset setting of Tabel.~\ref{tab:cross}).}
	\vspace{-12pt}
	\label{tab:lambda}
\end{table}
\subsection{Effects of Target Shape Compactness}

In this section, we investigate the impact of shape-awareness on the performance of SAZS in zero-shot semantic segmentation by analyzing the correlation between the mean intersection-over-union (mIoU) and the shape compactness (CO) of each category. Shape compactness, as proposed by Schick and others in 2012 \cite{schick2012measuring}, is a commonly used metric for measuring the similarity of superpixels to circles, which we use to characterize the shapes of objects in the input images.

For each input image in the PASCAL-${5}^i$ dataset, we collected the compactness (CO) metric of the ground-truth mask for the target object to describe its shape. We then calculated the variance of CO for each object category and plotted the results in Fig.~\ref{fig:CO}. The sample points in the figure represent the IoU and CO variance of each category, with the color indicating the experiment settings. This analysis aims to investigate how shape-awareness affects the SAZS's performance on zero-shot semantic segmentation.

The results demonstrate a negative correlation between the IoU and the CO variance of a specific category (with a Pearson correlation coefficient of $r > 0.7$ and $P<=0.001$), and the degree of correlation is higher for SAZS than for the baselines. These findings strongly suggest that shape-awareness can improve segmentation performance when objects have more stable shapes, and that SAZS is more able to leverage shape information compared to the other baselines. The experiments were conducted on the PASCAL-${5}^i$ dataset.

\subsection{Effects of the Language Embedding Locality}
Intuitively, distribution of language anchors in the latent feature space may largely affect vision-language alignment and thus the performance of the proposed method.
Inspired by recent research \cite{radford2021learning,kamath2021mdetr,li2020hero}, we model the distribution by the embedding locality of anchors which is defined by the mean value and standard deviation of euclidean distances in the feature space between one anchor and all other anchors. 

\begin{figure}[tbp]
  \begin{minipage}[t]{0.5\linewidth}
    \centering
    \includegraphics[height=4cm]{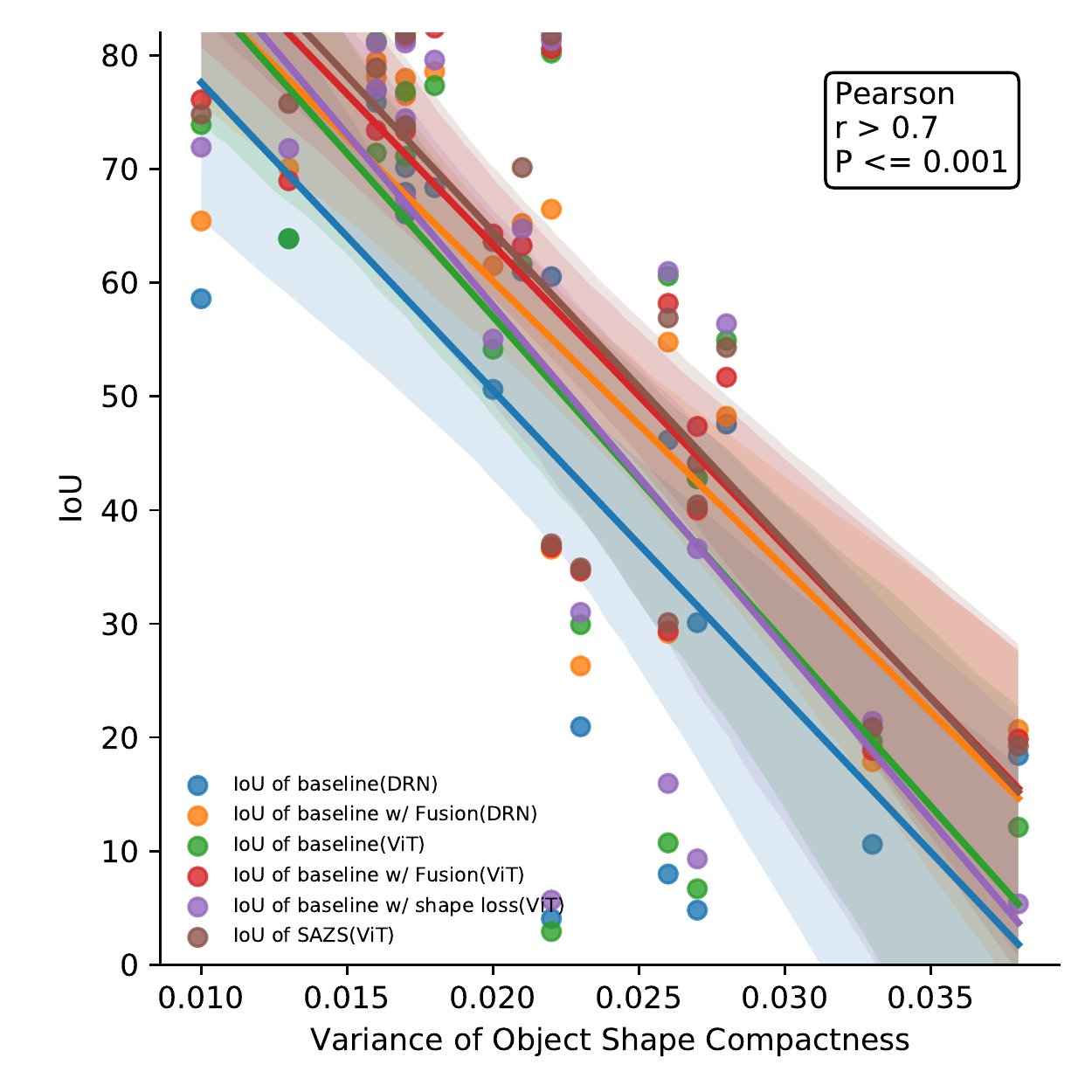}
    \subcaption{ }
    \label{fig:CO}
  \end{minipage}%
  \begin{minipage}[t]{0.5\linewidth}
    \centering
    \includegraphics[height=4cm]{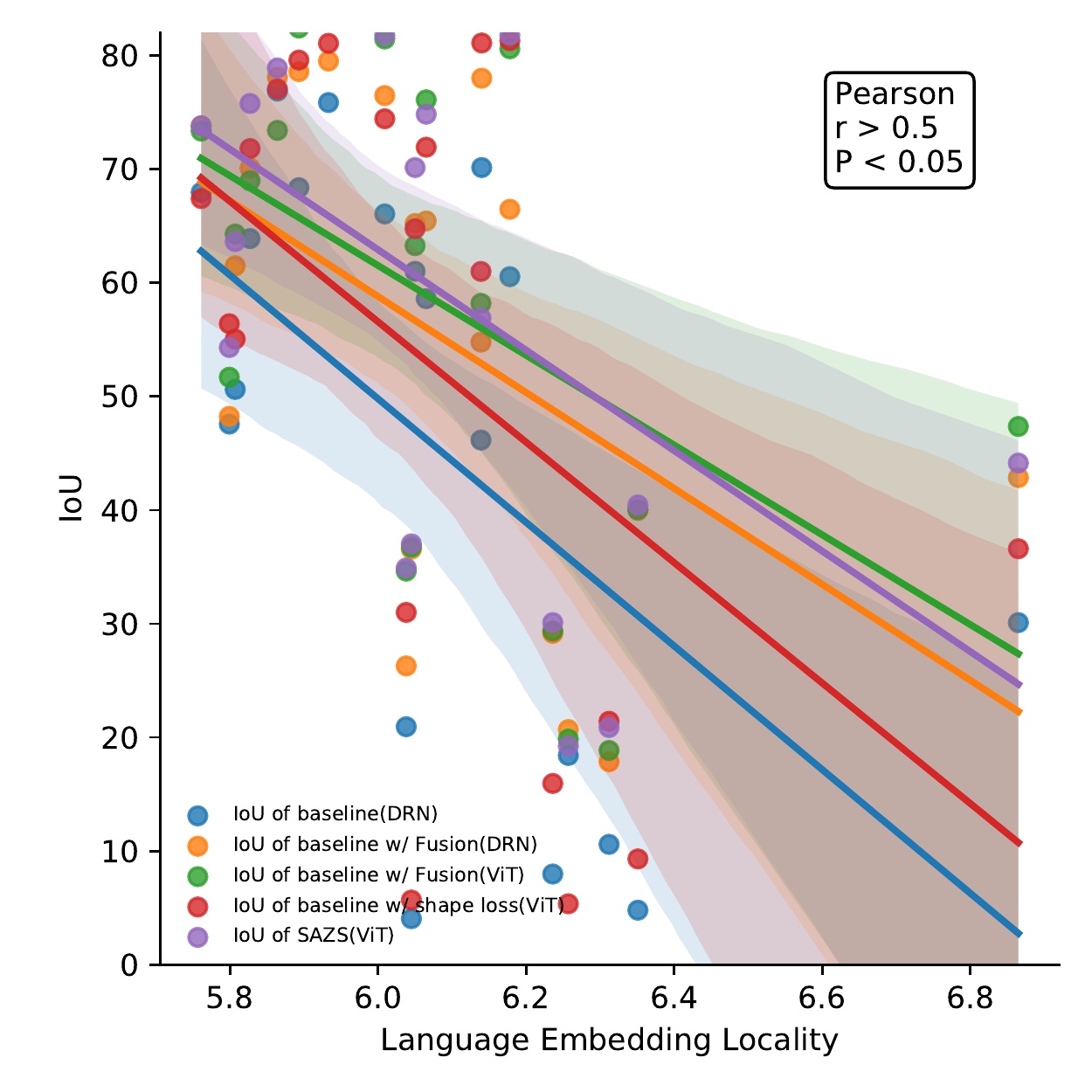}
    \subcaption{ }
    \label{fig:lang1}
  \end{minipage}
  \caption{Correlation of CO variance (a) or mean embedding locality (b) with IoU.}
  \label{fig:both}
\end{figure}

For each category in each setting of experiments, we calculate its embedding locality and report the results collected on PASCAL-${5}^i$ in Fig.~\ref{fig:lang1}.
The coordinates of sample points represent the IoU and the embedding locality of the corresponding category while the colors of the sample points denote the experiment settings.

According to the plotted results, we observe a negative linear correlation (with Pearson correlation coefficient $r > 0.5$ and $P \leq 0.05$) between the embedding locality mean and IoU of a certain category, indicating that the closer a category is in the feature space to the others, the easier it is for the visual and text embeddings which leads to higher performances.
Also, the degree of relevance of SAZS is the highest among all methods which implies that SAZS is able to better align pixel-wise visual embeddings towards the text anchors in the CLIP feature space.

\section{Conclusion}
In this paper, we present a novel framework for \textbf{S}hape-\textbf{A}ware \textbf{Z}ero-\textbf{S}hot semantic segmentation (abbreviated as \textbf{SAZS}).
The proposed framework leverages the rich priors contained in the feature space of a large-scale pre-trained visual-language model, while also incorporating shape-awareness through joint training on a boundary detection constraint task. This is necessary to compensate for the absence of fine-grained features in the feature space.
In addition, self-supervised spectral decomposition is used to obtain feature vectors for images, which are fused with the network predictions as prior knowledge to enhance the model's ability to perceive shapes.

Extensive experiments demonstrate the state-of-the-art performance of SAZS with significant margins over previous methods. Correlation analysis further highlights the impact of shape compactness and distribution of language anchors on the framework's performance. Our approach effectively exploits the shape of targets and feature priors, showing the highest correlation among all compared methods and proving the novelty of the shape-aware design.

\newpage

{\small
\bibliographystyle{ieee_fullname}
\bibliography{egbib}
}
\clearpage

\section{Appendix}

\subsection{Per-Category Evaluation}
\label{sec:intro}

Table \ref{tab:more_coco} and Table \ref{tab:more_pascal} demonstrate our per-category zero-shot semantic mIoU results on COCO-${20}^i$ \cite{lin2014microsoft} and PASCAL-${5}^i$ \cite{everingham2015pascal}, respectively. The mIoU of our proposed SAZS network structure demonstrates superior performance compared to the baseline. We also observed that certain categories often appear as small regions, such as ties, or have complicated internal structures, such as people. For these categories, textual feature guidance alone cannot provide sufficient information for semantic parsing, and the baseline without shape-awareness cannot effectively segment objects under self-supervision. However, when using a SAZS model, the mIoUs of these categories better align with the shapes of the objects than the baseline, which confirms that shape awareness indeed improves zero-shot learning.
\subsection{Speed and Complexity}
\label{sec:intro}

We conducted experiments to analyze the per-episode inference time and floating point operations per second (FLOPs) in order to demonstrate the complexity of our proposed approach. The results are summarized in Table \ref{tab:c_time} for the COCO-${20}^i$ dataset. Compared to the baseline model without the fusion module, SAZS had slower inference time, but significantly better performance. Although losses, including ${L}_{\rm shape}$, in our model did not add any time cost during inference, there is still potential for optimization in terms of inference speed and model complexity, which is exactly the direction for our future research.

\subsection{More Qualitative Results}

In this section, we provide additional qualitative results of our model with a ViT-L backbone on PASCAL-${5}^i$ and COCO-${20}^i$ datasets to demonstrate the model's ability to perform semantic segmentation on previously unseen categories.
Fig.~\ref{fig:pascal} showcases the results on PASCAL-${5}^i$, where all categories are unseen in their respective fold. The images presented in the figure vary in their content and complexity, and we display different visualizations of SAZS to demonstrate its versatility.

The results presented in Fig.\ref{fig:pascal} demonstrate the efficacy of SAZS in distinguishing the target semantic objects, such as bicycle, dining table, and TV monitor, from distractors like person, dog, and keyboard. 
Furthermore, in Fig.~\ref{fig:pascal}, SAZS accurately segments multiple instances of the target object, as is the case with the train, potted plant, and TV monitor.

Overall, these results demonstrate the robustness of our model in semantically segmenting novel categories with high precision and accuracy, even in complex scenes.
In this section, we present the visualization of COCO-${20}^i$ in Fig.\ref{fig:coco}, which includes both seen and unseen categories. We selected 20 scene and attribute labels with different semantics and multiple objects to demonstrate the versatility of SAZS. Despite the presence of noise and complexity in the scenes, SAZS accurately recognizes novel categories that are small and intricate, as illustrated by the examples of broccoli, pottedplant, and skis in Fig.\ref{fig:coco}.

In particular, in the second image of lines 2 and 3 of Figure 1, where multiple species appear in the scene with complex shapes, SAZS performs sharp object edge segmentation to accurately distinguish broccoli, carrots, and hot dogs.

Given the diversity of the presented scenes, we believe that SAZS is precise enough to be applied to various scenarios, including open scenario understanding and intelligent service robots.

\subsection{More Scatter Analysis}
\begin{figure*}[h]
    \centerline{
    \includegraphics[width=1\linewidth,height=0.7\linewidth]
    {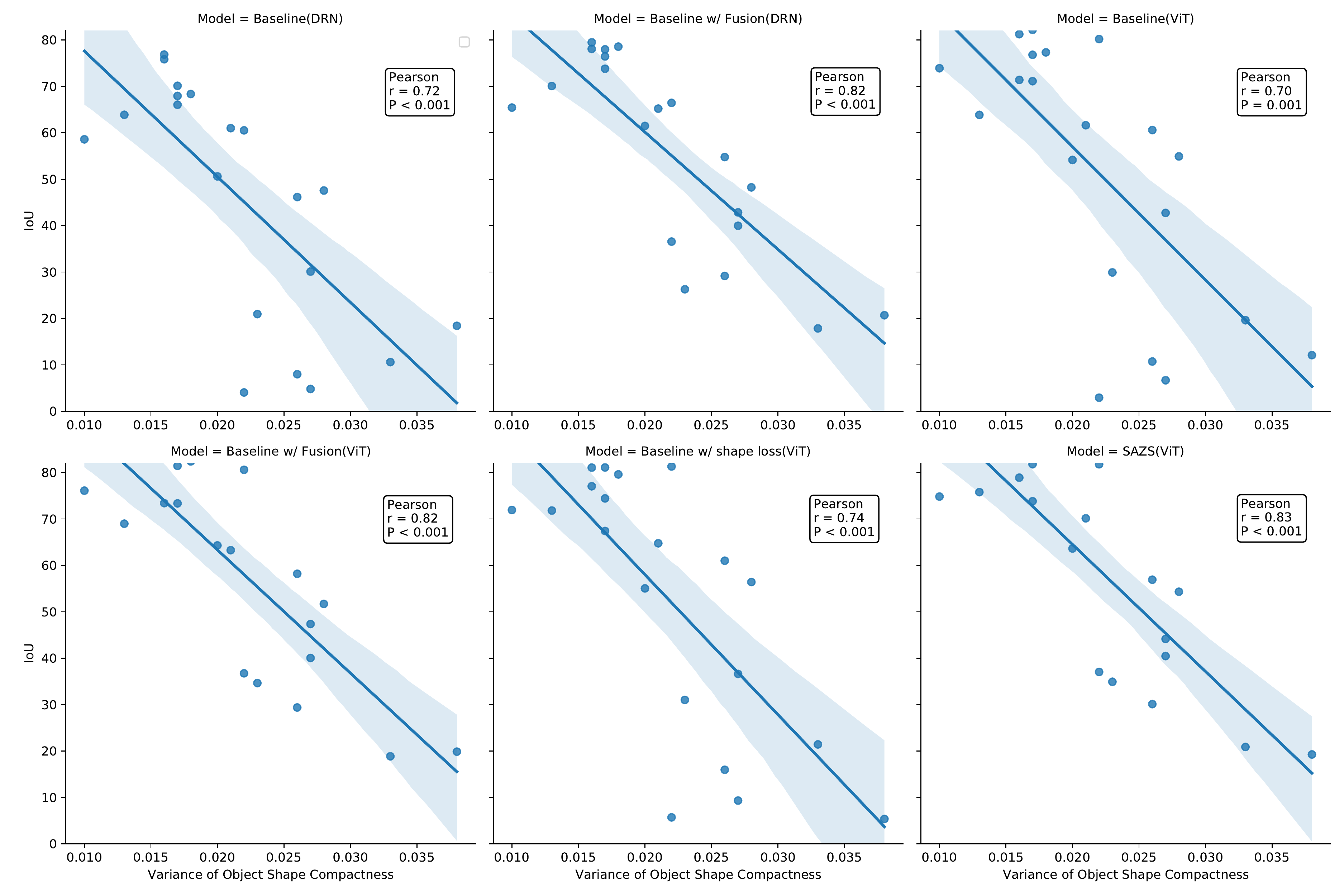}}
    \caption{More scatterplots on PASCAL-${5}^i$.}
    \label{fig:supp_co}
\end{figure*}

\begin{figure*}[h]
    \centerline{
    \includegraphics[width=0.96\linewidth]
    {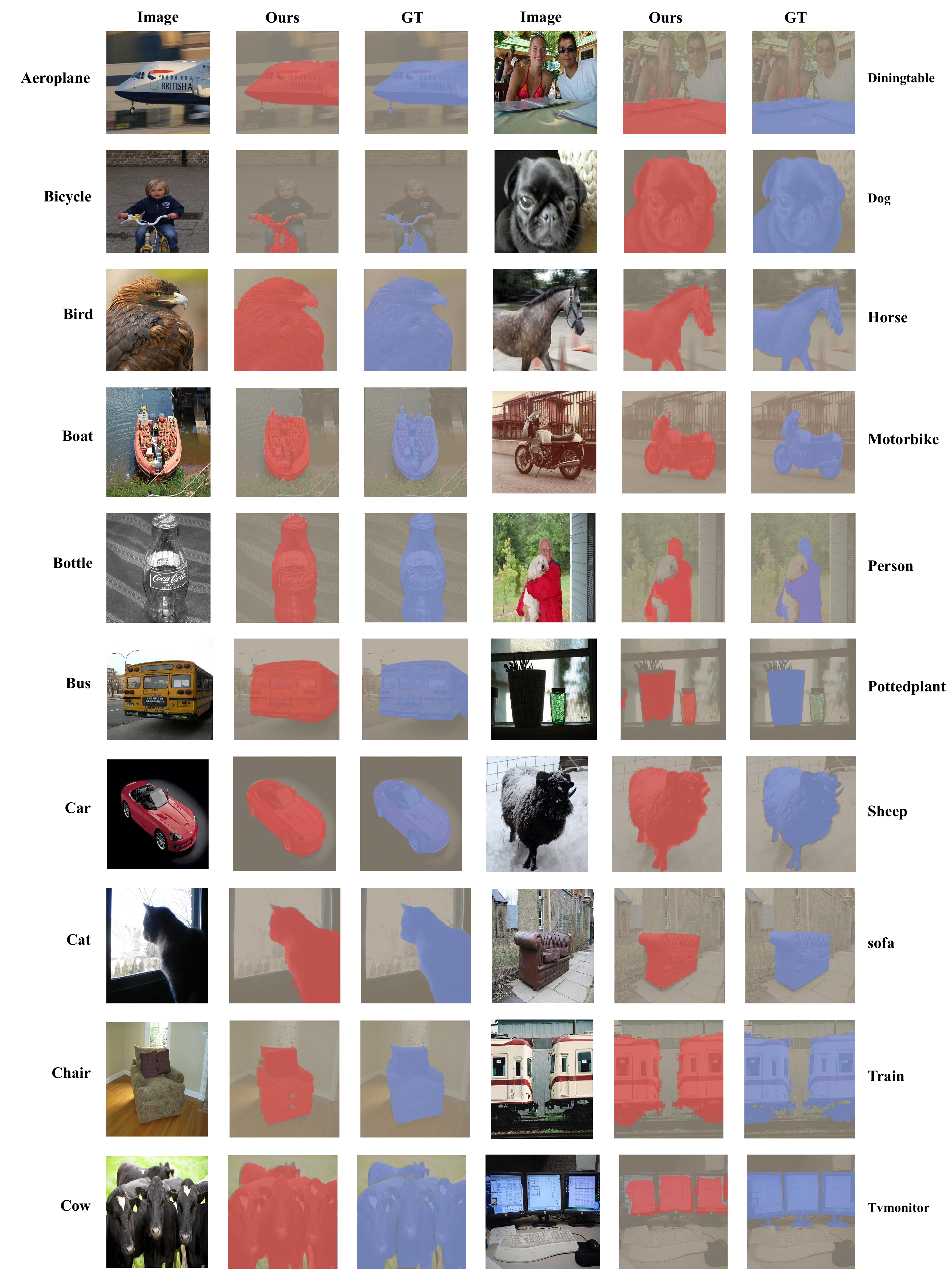}}
    \caption{More qualitative results on PASCAL-${5}^i$.}
    \label{fig:pascal}
\end{figure*}

\begin{figure*}[h]
    \centerline{
    \includegraphics[width=0.95\linewidth]
    {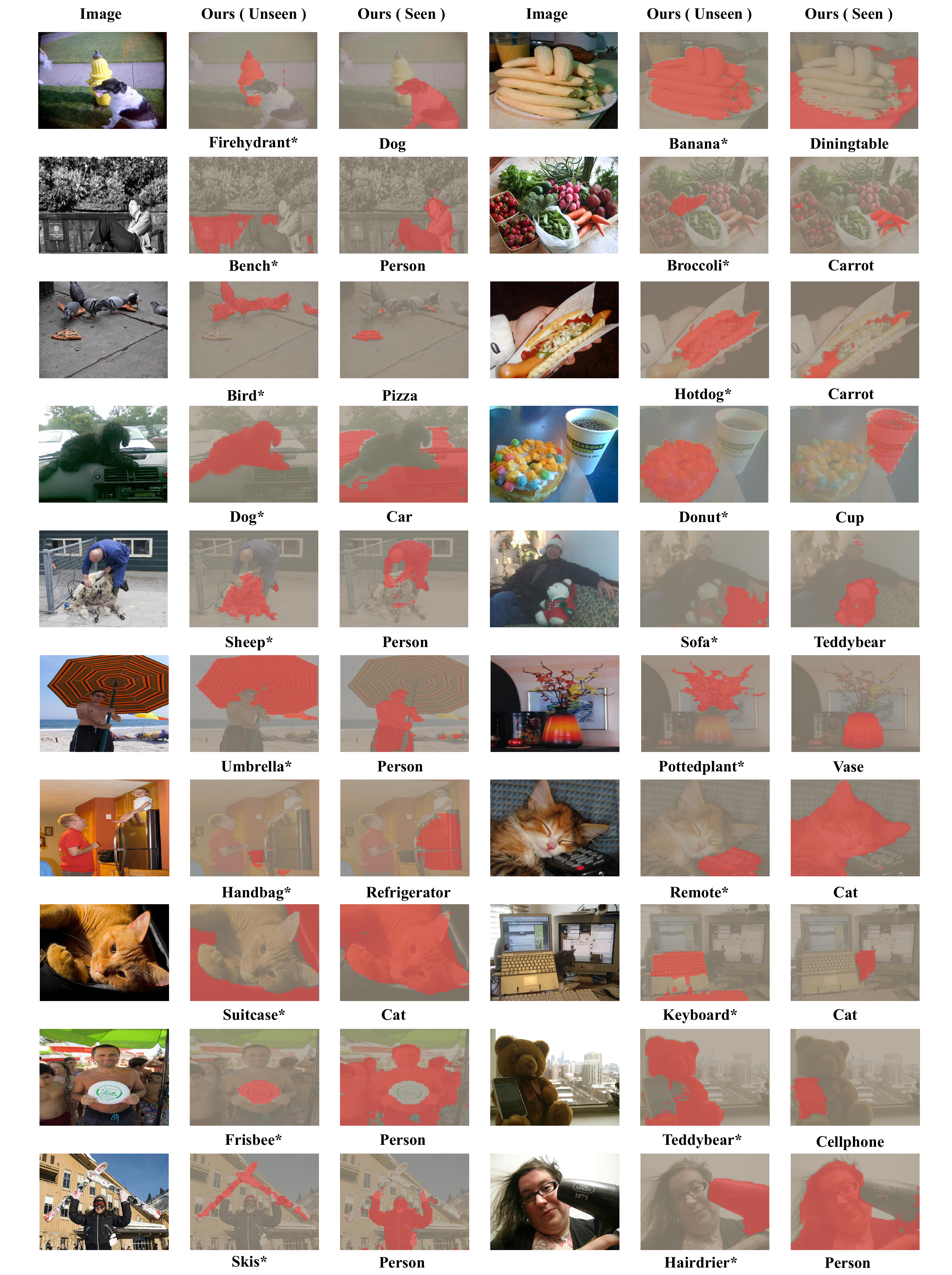}}
    \caption{More qualitative results on COCO-${20}^i$.}
    \label{fig:coco}
\end{figure*}

Fig.\ref{fig:supp_co} presents additional scatterplots and corresponding Pearson analysis results for the pascal dataset. The sample points in Fig.\ref{fig:supp_co} represent the IoU and CO variance of each model, and they demonstrate a negative correlation. The results indicate that our approaches, particularly those that incorporate shape-awareness, can increase the correlation between per-category IoU results and CO. For example, in the third column of Fig.~\ref{fig:supp_co}, the Pearson correlation coefficient $r$ of SAZS is 0.13 higher than that of the baseline.

\begin{table}
    \scriptsize
	\centering
	\begin{tabular}{c|c|ccc}
		\toprule
		\textbf{Model} & \textbf{Backbone} & \textbf{mIoU} & \ \textbf{time(s)} & \textbf{FLOPS(G)} \\
		\midrule
            w/o fusion & DRN & 26.6 & 177.43 &  275.76 \\
	    w/o fusion & ViT-L & 29.1 & 196.95 &  345.99 \\
            SAZS  & DRN & 35.2 & 230.54 &   275.76\\
            SAZS  & ViT-L & 35.3 & 222.52 &   345.99\\
		\bottomrule
	\end{tabular}
	\vspace{-4pt}
	\caption{More quantitative results on COCO-${20}^i$.}
	\vspace{-18pt}
	\label{tab:c_time}
\end{table}

\begin{table*}[htbp]
\centering
\caption{ Per-category zero-shot semantic segmentation results on COCO-${20}^i$.}
\renewcommand\arraystretch{1.8}{
\setlength\tabcolsep{0.3pt}{
\begin{tabular}{cc|cccccccccccccccccccc|c}
\hline \text { Method } &\text { Backbone } & \rotatebox{90} { person } &  \rotatebox{90} { Bicycle } &  \rotatebox{90}{ Car } & \rotatebox{90} { motorbike } &  \rotatebox{90} { aeroplane } &  \rotatebox{90} { Bus } &  \rotatebox{90}{ train } & \rotatebox{90}{ truck } &  \rotatebox{90} { boat } &  \rotatebox{90} { trafficlight } &  \rotatebox{90}{ firehydrant }&  \rotatebox{90}{ stopsign }&  \rotatebox{90}{ parkingmeter }&  \rotatebox{90}{ bench }&  \rotatebox{90}{ bird }&  \rotatebox{90}{ cat }&  \rotatebox{90}{ dog }&  \rotatebox{90}{ horse }&  \rotatebox{90}{ sheep }&  \rotatebox{90}{ cow }&  \rotatebox{90}{ mIoU } \\
\hline 
\text { Baseline } & \text {DRN} &35.7 &55.5 &38.2 &43.6 &69.2 &69.9 &15.1 &27.2 &\textbf{20.2}  &\textbf{24.0 }&12.2 & 5.9 &57.2 &66.5 &11.9 &43.3 &12.3 &\textbf{19.7} &25.3 &20.7 &35.2
 \\  
\textbf {Ours } & \text {DRN} & \textbf {36.0}
 & \textbf {61.5} & \textbf {38.5} & \textbf {55.7} & 66.7  & 72.2 & \textbf{17.9} & \textbf{29.4} & 16.7 & 14.4 & 12.2  & 5.9  & 53.8  & 65.8  & 11.5  & \textbf{46.2 } & 13.5  & 15.5  &\textbf{ 27.6 } & \textbf{22.5} &35.2
\\ 
\text {Baseline } & \text {ViT-L} & 35.7 & 55.1 & 32.1 & 47.2 & \textbf {75.6 } & \textbf{83.5} & 16.2  & 20.3 & 16.1  & 12.4 & 12.2 & 5.9 & \textbf{60.1} & \textbf{72.2} & 12.0 & 36.3 & 11.0 & 15.8 & 25.2 & 20.7 & 34.7\\
\textbf {Ours } & \text {ViT-L} & 35.7 & 56.5  & 33.4  & 48.2   & 74.7  & 83.2  & 16.2  & 25.0  & 17.6   & 13.1  & 12.1 & \textbf{7.3 }&56.4 &71.9 &\textbf{12.3 }&35.3 &\textbf{13.8} &17.6 &25.3 &21.1 &\textbf{35.3}
\\
\hline \text { Method } & \text { Backbone }  & \rotatebox{90} { elephant } & \rotatebox{90} { bear } & \rotatebox{90} { zebra } & \rotatebox{90} { giraffe } & \rotatebox{90} { backpack } & \rotatebox{90} { umbrella } & \rotatebox{90} { handbag } & \rotatebox{90} { tie } & \rotatebox{90} { suitcase } & \rotatebox{90} { frisbee } & \rotatebox{90} { skis }& \rotatebox{90} { snowboard }& \rotatebox{90} { sportsball }& \rotatebox{90} { kite }& \rotatebox{90} { baseballbat }& \rotatebox{90} { baseballglove }& \rotatebox{90} { skateboard }& \rotatebox{90} { surfboard }& \rotatebox{90} { tennisracket }& \rotatebox{90} { bottle }&  \\
\hline
\text { Baseline } & \text { DRN }&\textbf{25.6} &\textbf{67.0 }&\textbf{23.9} &16.5 &\textbf{64.6 }&74.5 &35.5&27.5 &\textbf{55.3} &\textbf{49.7} &10.3 &21.1 &41.0 &78.8 &\textbf{28.7} &33.5 &18.8 &18.2 &12.1 & 60.5 
\\ 

 \textbf{Ours } &  \text { DRN } & 24.3  & 63.8 & 23.2 & 16.5 & 57.0 & 74.5 & \textbf{35.8} &38.2 &53.3 &40.6 &\textbf{10.9}&21.1 & 38.5 &63.9 &20.9 &30.3 &20.3 &18.2 &\textbf{15.9} &62.2 & \\
 \text { Baseline } & \text {ViT-L} &  15.9 &  61.3 &  18.8 & \textbf{16.9 }& 60.0  & \textbf{79.0 }& 35.5 & 52.1 & 51.9 &47.0 &10.4 &21.1 &\textbf{43.7 }&\textbf{85.6 }&20.1 
&\textbf{33.9} 
&\textbf{24.7 }
&18.9 
&13.0 
&\textbf{69.9 }
&\\
\textbf { Ours } &  \text {ViT-L} & 16.2 & 57.1 & 19.3  & 16.5  & 61.0  & 78.7  & 35.5 & \textbf{53.7}  & 52.1 & 43.6  & 10.3 &21.1 &40.8 &78.5 &19.8 &32.8 &21.4 &\textbf{18.9 }&15.0 &69.2 & \\
\hline \text { Method } & \text { Backbone }  & \rotatebox{90} { wineglass} & \rotatebox{90} { cup} & \rotatebox{90} { fork} & \rotatebox{90} { knife} & \rotatebox{90} { spoon} & \rotatebox{90} { bowl} & \rotatebox{90} { banana} & \rotatebox{90} { apple} & \rotatebox{90} { sandwich} & \rotatebox{90} { orange} & \rotatebox{90} { broccoli}& \rotatebox{90} { carrot}& \rotatebox{90} { hotdog}& \rotatebox{90} { pizza}& \rotatebox{90} { donut}& \rotatebox{90} { cake}& \rotatebox{90} { chair}& \rotatebox{90} { sofa}& \rotatebox{90} { pottedplant}&  \rotatebox{90} { bed} \\
\hline 
\text { Baseline } & \text { DRN } &\textbf{16.8 }
&56.2 
&\textbf{60.3 }
&\textbf{47.3 }
&72.5 
&73.9 
&4.7 
&9.8 
&9.6 
&18.8 
&3.7 
&\textbf{46.7 }
&\textbf{38.1 }
&62.2 
&10.2 
&17.2 
&28.1 
&\textbf{13.0} 
&42.0 
&36.7 
\\
 \textbf{ Ours } &  \text { DRN } &  15.0 & 57.0 & 49.6 & 44.8 & 73.5 & 77.1 & \textbf{5.0} & 8.6 & 9.7 &\textbf{23.4} & 3.9 &46.4 &37.4 &\textbf{67.7} &10.5 &17.2 &37.0 &12.0 &\textbf{49.1} &47.0 &\\
 \text { Baseline } & \text {ViT-L} &15.4 
  &\textbf{62.5 }
 & 52.3 
  & 38.8 
  & 78.8 
  & 79.5 
  & 4.8 
 & 8.3 
  &\textbf{ 9.8 }
 &16.1 
&3.7 
&42.9 
&37.4 
&60.0 
&10.5 
&\textbf{25.5 }
&\textbf{39.6 }
&11.6 
&39.1 
&41.5 
& \\
\textbf { Ours } &  \text {ViT-L} & 14.5   & 58.6 
  & 58.9 
  & 39.0 
 & \textbf{79.3 }
 & \textbf{80.2 }
  & 4.8 
  &\textbf{10.2 }
  & 9.7 
 & 16.0 
 & \textbf{4.1}
 &44.6 
 &37.4 
 &60.6 
 &10.5 
 &18.1 
 &36.7 
 &11.4 
 &46.7 
 &\textbf{47.3 }
 & \\

\hline \text { Method } & \text { Backbone }  &  \rotatebox{90} { diningtable} & \rotatebox{90} { toilet} & \rotatebox{90} { tvmonitor} & \rotatebox{90} { laptop} & \rotatebox{90} { mouse} & \rotatebox{90} { remote} & \rotatebox{90} { keyboard} & \rotatebox{90} { cellphone} & \rotatebox{90} { microwave } & \rotatebox{90} { oven}& \rotatebox{90} { toaster}& \rotatebox{90} { sink}& \rotatebox{90} { refrigerator}& \rotatebox{90} { book}& \rotatebox{90} { clock}& \rotatebox{90} { vase}& \rotatebox{90} { scissors}& \rotatebox{90} { teddybear}& \rotatebox{90} { hairdrier}& \rotatebox{90} { toothbrush
} \\
\hline 
\text { Baseline } & \text { DRN } &39.6&\textbf{55.9 }&44.6 & 74.3 &77.7 &70.6  &14.0  & 32.5 & 13.1 & \textbf{12.0} &\textbf{8.1 }&25.9 
&24.0 
&34.3 
&\textbf{43.7} 
&25.7 
&37.3 
&11.4 
&30.9 
& 33.5 &
\\
 \textbf { Ours } &  \text { DRN } & \textbf{54.7 }
  & 44.9 
 & 47.6 
 & 60.5 
 & 70.5 
& 64.0 
 &\textbf{16.7 }
 &\textbf{34.5 }
 &\textbf{13.8 }
 &11.2 
&7.2 
 &26.2 
 &24.8 
 &\textbf{42.6} 
 &43.6 
 &26.0 
 &\textbf{47.6 }
&\textbf{15.5 }
 &\textbf{32.6 }
 &\textbf{28.4 }
& \\
 \text { Baseline } & \text {ViT-L} &  39.2 
 &32.6 
   &45.7 
   & 70.4 
 & \textbf{79.5 }
  & 67.0 
  & 14.0 
 & 20.6 
  &13.4 
  & 10.7 
&6.4 
&26.1 
&24.1 
&37.7 
&41.7 
&25.7 
&32.6 
&11.9 
&27.4 
&26.8 
 \\
\textbf { Ours } &  \text {ViT-L} & 43.9 
 & 38.2 
  & \textbf{53.1 }
  & \textbf{77.0 }
  & 78.9 
  &\textbf{71.7 }
  & 14.0 
  & 16.4 
  & 13.1 
 & 10.8 
 & 6.2 
 &\textbf{26.9 }
 &\textbf{27.9 }
 &40.8 
 &41.8 
 &\textbf{26.8 }
 &41.9 
 &12.1 
 &29.8 
 &28.1 
 & \\
\hline
\end{tabular}}}
\label{tab:more_coco}%
\end{table*}%
\begin{table*}[htbp]
\centering
\caption{ Per-category zero-shot semantic segmentation results on PASCAL-${5}^i$.}
\renewcommand\arraystretch{1.2}{
\setlength\tabcolsep{5pt}{
\begin{tabular}{cc|cccccccccc|cc}
\hline \text { Method } &\text { Backbone } & \rotatebox{90} { Aeroplane } &  \rotatebox{90} { Bicycle } &  \rotatebox{90}{ Bird } & \rotatebox{90} { Boat } &  \rotatebox{90} { Bottle } &  \rotatebox{90} { Bus } &  \rotatebox{90}{ Car } & \rotatebox{90}{ Cat } &  \rotatebox{90} { Chair } &  \rotatebox{90} { Cow } &  \rotatebox{90}{ mIoU }&  \rotatebox{90}{ FBIoU } \\
\hline 
\text { Baseline } & \text {DRN} &58.6 
 & 20.9 
 & 68.4 
 & 50.6
 & 46.2
 & 76.9
 & 47.6 
 & 70.1
 & 10.6
 & 75.9 
 & 45.5
 & 61.7
 \\  
\textbf {Ours } & \text {DRN} & 65.4
 & 26.3
 & 78.6
 & 61.5
 & 54.8
 & 78.1
 & 48.3
 & 78.0
 & 17.9
 & 79.5
 & 55.5
 & 66.4\\ 
\text {Baseline } & \text {ViT/L} & \textbf {76.1}  & 34.6  & 82.4  & \textbf{64.3}  & \textbf{58.2}  & 73.4  & 51.7  & \textbf{84.7}  & 18.9  & 83.1  & 58.4 &68.3 \\
\textbf {Ours } & \text {ViT/L} & 74.8 & \textbf {34.9} & \textbf {83.0}  & 63.6  & 56.9 & \textbf{78.9}  & \textbf{54.3} & 84.0 & \textbf{20.9}  & \textbf{83.2}  & \textbf{59.4} & \textbf{69.0}\\
\hline \text { Method } & \text { Backbone }  & \rotatebox{90} { Diningtable } & \rotatebox{90} { Dog } & \rotatebox{90} { Horse } & \rotatebox{90} { Motorbike } & \rotatebox{90} { Person } & \rotatebox{90} { Pottedplant } & \rotatebox{90} { Sheep } & \rotatebox{90} { Sofa } & \rotatebox{90} { Train } & \rotatebox{90} { Tvmonitor } & & \\
\hline 
\text { Baseline } & \text { DRN } & 4.8
  & 66.1   & 68.0  & 61.0  & 4.1  &  18.4  & 60.5  & 30.1  & 63.9
 & 8.0 &\\
 \textbf { Ours } &  \text { DRN } & 40.0 & 76.5  & 73.8 &  65.2 & 36.6 &  \textbf{20.7} & 66.5  & 42.9  & 70.1  & 29.2 &  \\
 \text { Baseline } & \text {ViT/L} & 40.0  & 81.5  & 73.4  & 63.3  & 36.7  & 19.9  & 80.6  & \textbf{47.4}  & 69.0 & 29.4 \\
\textbf { Ours } &  \text {ViT/L} & \textbf{40.5}  & \textbf{81.8}  & \textbf{73.8}  & \textbf{70.1}  & \textbf{37.0}  & 19.3 & \textbf{81.8}  & 44.1 & \textbf{75.8} & \textbf{ 30.1 } &  \\
\hline
\end{tabular}}}
\label{tab:more_pascal}%
\end{table*}%
\end{document}